\documentclass[runningheads]{llncs}
\usepackage{graphicx}
\begin{document}
\title{Towards A Sentiment Analyzer for Low-Resource Languages}
%
\author{Dian Indriani\inst{1} \and
Arbi Haza Nasution\inst{1}\and
Winda Monika\inst{2} \and
Salhazan Nasution\inst{3}}
\authorrunning{Indriani et al.}
\institute{Informatics Engineering, Universitas Islam Riau, Indonesia \and
	Library Science, Universitas Lancang Kuning, Indonesia \and
	Informatics Engineering, University of Riau, Indonesia \\
	\email{diianindrianii23@student.uir.ac.id, arbi@eng.uir.ac.id, windabi.wm@gmail.com, salhazan@lecturer.unri.ac.id}}
\maketitle              
\begin{abstract}
Twitter is one of the top influenced social media which has a million number of active users. It is commonly used for microblogging that allows users to share messages, ideas, thoughts and many more. Thus, millions interaction such as short messages or tweets are flowing around among the twitter users discussing various topics that has been happening world-wide. This research aims to analyse a sentiment of the users towards a particular trending topic that has been actively and massively discussed at that time. We chose a hashtag \textit{\#kpujangancurang} that was the trending topic during the Indonesia presidential election in 2019. We use the hashtag to obtain a set of data from Twitter to analyse and investigate further the positive or the negative sentiment of the users from their tweets. This research utilizes rapid miner tool to generate the twitter data and comparing Naive Bayes, K-Nearest Neighbor, Decision Tree, and Multi-Layer Perceptron classification methods to classify the sentiment of the twitter data. There are overall 200 labeled data in this experiment. Overall, Naive Bayes and Multi-Layer Perceptron classification outperformed the other two methods on 11 experiments with different size of training-testing data split. The two classifiers are potential to be used in creating sentiment analyzer for low-resource languages with small corpus.

\end{abstract}

\section{Introduction}
A rich sentiment analysis corpus is crucial in creating a good sentiment analyzer. Unfortunately, low-resource languages like Indonesian lack such resources. Some prior studies focused on enriching low-resource languages \cite{nasution2018PHMT,nasution2016pivot,nasution2017pivot,nasution2017plan,nasution2017simcluster,nasution2018collab,nasution2019simcluster,nasution2017PHMT}. 
The rapid growth of online textual data creates an urgent need for powerful text mining techniques \cite{aggarwal2012mining}. Sentiment analysis or opinion mining is a part of text mining. Sentimen analysis basically is a computational research that analyses the textual expression from opinion, sentiment and emotion of the social media users \cite{liu2012sentiment}. It extracts attributes and components of the documented object. Through the sentiment analysis of the text, information such as the public's emotional status, views on some social phenomena, and preferences for a product can be obtained \cite{Yuan2019}. Hence, the perspective of the users either positive or negative could be revealed. 

During the Indonesia 2019 presidential election, the competition was quite fierce where there were only two candidates fighting in the battle. Most of supporters from these two candidates were actively campaigning their candidates on social media and twitter was the highly used social media chosen by them. Due to the huge enthusiasm of those two supporters, most of the time fierce debate among them could not be avoided.   
One of the trending topic emerged was during the recapitulation of the votes. Twitter users reacted to the several findings showed that the calculation of the votes led to deception. Foremost, supporters from one party, from Prabowo Subianto volunteers found that many evidence of the wrong data were inputed to the system. Thus, the real count results was irrelevant with the information displayed on the system. This finding made the situation in Indonesia heating up. Supporters from Prabowo Subianto was upset and condemned the General Election Commision as the legal institution to take full responsibility of this matter. To express their disappointment, most of the twitter users created hashtag \textit{\#kpujangancurang} or ``The General Election Commision should not be unfair". However, this issue was objected by the opponent supporters. They argued that this issue was merely caused by human error. The same hashtag actually was being used by the both parties, so that no one knows the exact sentiment of the tweets. Therefore, sentiment analyzer that could analyse the sentiment of the tweets is crucial

In sentiment analysis, the available corpus in Indonesian language is scarce. The existing machine learning tool such as rapidminer has two sentiment analyzer which are Aylien and Rosette, do not cover Indonesian language. We run an experiment by using the \textit{\#kpujangancurang} hastag to obtain corpus using rapidminer to extract the tweets and then analyse the sentiment of users by using four machine learning methods which are Naive Bayes, K-Nearest Neighbor, Decision Tree, and Multi-Layer Perceptron classification. The objective of this research is to find out which classifier is more suitable to be used in creating sentiment analyzer for low-resource languages with small corpus.

\section{Literature Study}
Several researches have been done on sentiment analysis. A study attempted to analyze the online sentiment changes of social media users using both the textual and visual content by analysing sentiment of twitter text and image \cite{You2016}.  Another related study performed linguistic analysis of the collected corpus and explain discovered phenomena to build a sentiment classifier, that is able to determine positive, negative and neutral sentiments for a document \cite{pak2010twitter}.

Furthermore, several studies have been done using machine learning method on sentiment analysis, for instance a study showed that a similar research on a twitter sentiment analysis by applying Naive Bayes classifier method to investigate the sentiment analysis of the twitter users on the traffic jam in Bandung \cite{rodiyansyah2012klasifikasi}. Another study focused on data classification using k-NN (k-Nearest Neighbors) and Naive Bayes where the Corpus was downloaded from TREC Legal Track with a total of more than three thousand text documents and over twenty types of classifications \cite{Rasjid2017}. A study utilized maximum entropy part of speech tagging and support vector machine to analyse the public sentiment. The study used dataset in Indonesian language and implemented machine learning approached due to its efficiency for integrating a large scale feature into a model. This kind of approach has been successfully implemented in various tasks such as natural language processing \cite{putranti2014analisis}. A study proposed a semi-automatic, complementary approach in which rule-based classification, supervised learning and machine learning are combined into a new method to achieve a good level of effectiveness \cite{prabowo2009sentiment}. Another study about opinion mining for hotel rating through reviews using decision tree classifier shows the advantage of using the algorithm is that the rule set can be easily generated and by analyzing each level of the tree, a particular service quality can be improved \cite{gupta2018opinion}. Deep learning methods also have been widely used in sentiment analysis tasks \cite{mukherjee2019malignant,zhang2018deep}. However, these studies show different accuracy from each machine learning method used depending on the size of the corpus. 

RapidMiner is an open source software\footnote{https://rapidminer.com}. RapidMiner is one of the solutions for doing analysis on data mining, text mining and prediction analysis. RapidMiner uses various descriptive technique and prediction in giving a new insight to the users so that allows and helps users to make a better decision. RapidMiner is a standalone software and enable to be integrated with its own products. RapidMiner provides GUI (Graphic User Interface) for designing an analytical pipeline. GUI will generate XML (Extensible Markup Language) that defines the analytical process of the users need to be applied on the data. This file is later on read by rapid miner to be automatically analyzed. We use rapid miner due to several reasons: it eases in getting the dataset, it can be connected to twitter, it enables to search the topic as query so that the intended topic will emerge and can be saved in excel file, furthermore it allows extracting plentiful data. A study examined an anomaly detection extension for RapidMiner in order to assist non-experts with applying eight different k-nearest-neighbor and clustering based algorithms on their data \cite{Amer2012}. However, in this study, we only use RapidMiner to extract data from Twitter.

\section{Research Methodology}
In this study, we use a dataset that was gotten from the tweets' document. We utilized rapid miner to obtain the tweets from the hashtag \textit{\#kpujangancurang}. To investigate further about the hashtag \textit{\#kpujangancurang}, we compare Naive Bayes, K-Nearest Neighbor, Decision Tree, and Multi-Layer Perceptron classification methods to classify the sentiment of the twitter data. There are two steps of the document classification: the first one is training the document that has been categorized. And the second one is training the uncategorized document. The four methods classify the distribution of the positive and negative sentiments. There are overall 200 labeled data in this experiment. To evaluate the performance of the sentiment analyzer, we use accuracy as the evaluation measure.

\subsection{System Workflow}
\subsubsection{Overview}
Sentiment analysis overview is described in details which is depicted in the Fig. 1 below. 

\begin{figure}[htbp]
	\centerline{\includegraphics[scale=0.33]{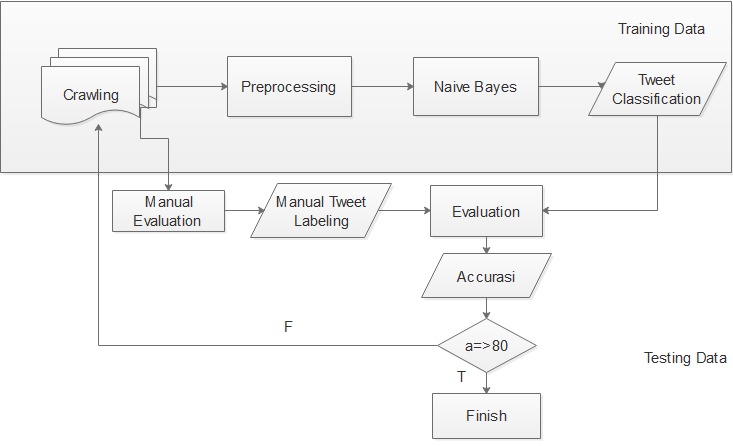}}
	\caption{Example of sentiment analysis workflow using Naive Bayes method.}
	\label{fig1}
\end{figure}

\begin{itemize}
	\item Data Crawling: It is a process of aggregating data from twitter using rapid miner as a tool. The aggregated data from hashtag \textit{\#kpujangancurang} is used as training dataset and testing dataset.
	\item Preprocessing: It is a process of cleaning the data by deleting common words by referring to stopwords. 
	\item Classification: Naive Bayes method is applied to classify the sentiment into positive and negative sentiments. The rest of methods will be used in the same manner.
	\item Evaluation: The classification result from classifiers is evaluated with the manual labeling classification. The accuracy of the classification determine whether a new training dataset need to be added or not to reach the accuracy threshold of 80\%.
\end{itemize}

\subsubsection{Dataset}
How do we get the dataset is depicted in Fig. 2 below. The dataset that we analyse is in Indonesian language. Firstly, the tweet was queried by using the hashtag \textit{\#kpujangancurang}.

\begin{figure}[htbp]
	\centerline{\includegraphics[scale=0.43]{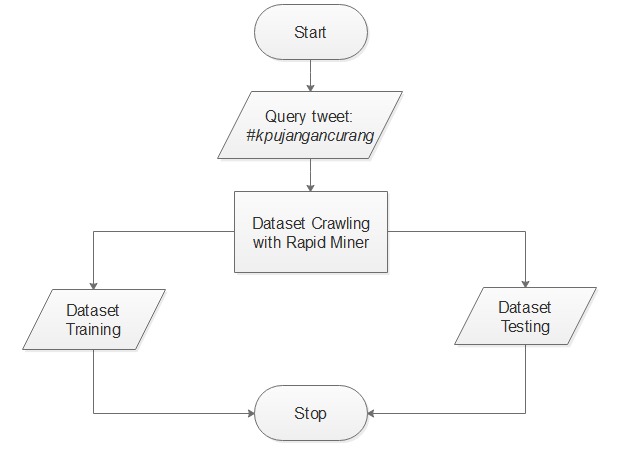}}
	\caption{Dataset Flow.}
	\label{fig2}
\end{figure}

Then, the queried data is crawled by using rapidminer. The result from the query is divided into two part: training data and testing data. Testing data is classified by using the classifiers and then the result was marked with negative and positive sentiment label. Whereas, the training data is classified manually and the result was marked the same way as testing data is treated. Training data will be used during the evaluation to determine the accuracy of the result. Table 1 shows example of evaluation of the predicted sentiment by the classifiers.

\begin{table}[htbp]
	\caption{Example of Evaluation of The Predicted Sentiment}
	\begin{center}
		\begin{tabular}{|p{5cm}|p{2cm}|p{2cm}|p{2cm}|}
			\hline
			\textbf{Testing Data}&\textbf{Predicted Sentiment}&\textbf{Manually Labeled Sentiment}&\textbf{Accuracy} \\
			\hline
			kalau terus melanggar, hukumannya segera diterapkan&Positive&Positive&Accurate\\
			kalau bersih kenapa takut audit forensic&Negative&Negative&Accurate\\
			harus banyak belajar ke @BKNgoid dalam hal penyelenggaraan akbar&Positive&Positive&Accurate\\
			Kebenaran meninggikan derajat bangsa tetapi dosa adalah noda bangsa&Negative&Positive&Inaccurate\\			
			\hline
		\end{tabular}
		\label{tab1}
	\end{center}
\end{table}

\subsubsection{Preprocessing}
Preprocessing process is an important step for the next step which disposes the non-useful attribute that can be noise for the classification process. Data that is imported in this process is a raw data, thus the result of this process is a high-quality document expected to ease the classification process. Preprocessing process is depicted in Fig. 3. 

\begin{figure}[htbp]
	\centerline{\includegraphics[scale=0.33]{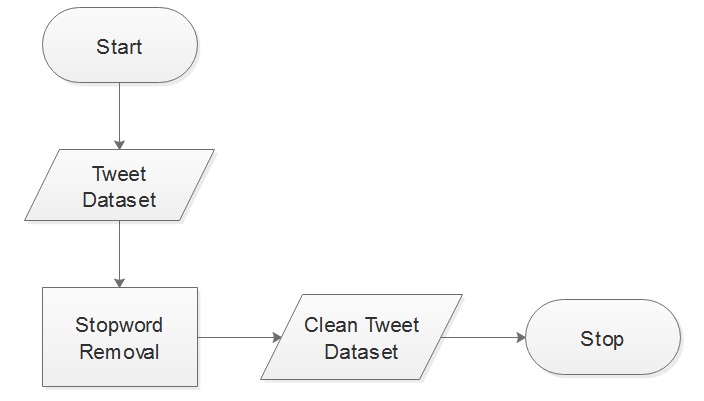}}
	\caption{Preprocessing Flow.}
	\label{fig3}
\end{figure}

\begin{table}[htbp]
	\caption{Preprocessing Process Example}
	\begin{center}
		\begin{tabular}{|p{1cm}|p{5cm}|p{5cm}|}
			\hline
			\textbf{Stage}&\textbf{Before}&\textbf{After} \\
			\hline
			1&Benar juga, kpu yang membuat rakyat resah. Aduh kejamnya kecurangan.&Benar juga kpu yang membuat rakyat resah Aduh kejamnya kecurangan\\
			
			2&Benar juga kpu yang membuat rakyat resah Aduh kejamnya kecurangan&benar juga kpu yang membuat rakyat resah aduh kejamnya kecurangan\\
			
			3&benar juga kpu yang membuat rakyat resah aduh kejamnya kecurangan&-benar- -juga- -kpu- -yang- -membuat- -rakyat- -resah- -aduh- -kejamnya- -kecurangan-\\
			
			4&-benar- -juga- -kpu- -yang- -membuat- -rakyat- -resah- -aduh- -kejamnya- -kecurangan-&-benar- -kpu- -membuat- -rakyat- -resah- -kejamnya- -kecurangan- \\
			\hline
		\end{tabular}
		\label{tab1}
	\end{center}
\end{table}

This step is started with punctuation removal, case folding, tokenizing, and finally stopword removal which is intended to remove words that are not relevant with the topic. If in the tweet document exists irrelevant words, then these words will be removed. An example of each stage of the preprocessing process is listed in Table 2. The detailed preprocessing stage is as follow:
\begin{itemize}
	\item Removing punctuation. This stage is the initial process in order to get pure text containing words only so that further processing becomes easier.
	\item Case Folding. This stage is the process of changing uppercase letters to lowercase letters.
	\item Tokenizing. In this stage, each word will be separated based on the specified space.
	\item Filtering. This stage is the removal of unimportant words based on Indonesian stopwords.
\end{itemize}

\subsubsection{Term Frequency - Inverse Document Frequency (TF-IDF)}
After doing the preprocessing, the next step is to weight the words using the tf-idf calculation. Tf-idf is a way of giving the weight of a word (term) to words. For single words, each sentence is considered as a document. The following is an example of tf-idf calculation. The example of documents that want to be weighted is shown in Table 3 and the sample tf-idf result of Document A is shown in Table 4.

\begin{table}[htbp]
	\caption{Example of Documents}
	\begin{center}
		\begin{tabular}{|l|p{7cm}|}
			\hline
			\textbf{Tweet Document}&\textbf{Text}\\
			\hline
			Document A&Jangan ancam rakyat, rakyat indonesia pintar\\

			Document B&Rakyat tidak pernah gagal bernegara, pemerintah yang gagal bernegara\\

			Document C&Suara rakyat dicuri, bagaimana uang rakyat\\
			\hline
		\end{tabular}
		\label{tab1}
	\end{center}
\end{table}

\begin{table}[htbp]
	\caption{TF-IDF Score of Document A}
	\begin{center}
		\begin{tabular}{|l|l|l|l|}
			\hline
			\textbf{Word}&\textbf{TF}&\textbf{IDF}&\textbf{Weight}\\
			\hline
			ancam &1&0.477&0.477\\

			bernegara&0&0.176&0\\

			gagal&0&0.176&0\\

			jangan&1&0.477&0.477\\

		    rakyat &0.4&-0.2218&-0.0887\\

			indonesia&1&0.477&0.477\\

			pintar&1&0.477&0.477\\

			tidak&0&0.477&0\\

			pernah&0&0.477&0\\

			pemerintah&0&0.477&0\\

			dicuri&0&0.477&0\\

			bagaimana&0&0.477&0\\

			uang&0&0.477&0\\
			\hline
		\end{tabular}
		\label{tab1}
	\end{center}
\end{table}
\subsubsection{Classifier}
The last step is classifying the weighted data with Naive Bayes, K-Nearest Neighbor, Decision Tree, and Multi-Layer Perceptron classification methods. To evaluate which classifiers are best for scarce corpus, we experimented by changing the size of the training-testing data split from 0.25-0.75 to 0.75-0.25. The evaluation is done by measuring the accuracy of the classifiers for each scenario as shown in Fig. 4.

\section{Result}
We obtained 200 twitter data using rapidminer. From the 200 twitter data, we conducted 11 experiments with different size of training-testing data split. Every classifiers shows a trend of increased accuracy on larger size of training data. However, Naive Bayes and Multi-Layer Perceptron classifier outperformed the other two methods in overall experiment as shown in Fig. 4. Decision Tree classifier shows a very low performance on small data, while K-Nearest Neighbor classifier shows accuracy below 0.76 on all combination size of training-testing data split. Both Naive Bayes and Multi-Layer Perceptron classifier have the highest accuracy on all combination size of training-testing data split and show consistent increased of accuracy as the training data size is increased.

\begin{figure}[htbp]
	\centerline{\includegraphics[scale=0.43]{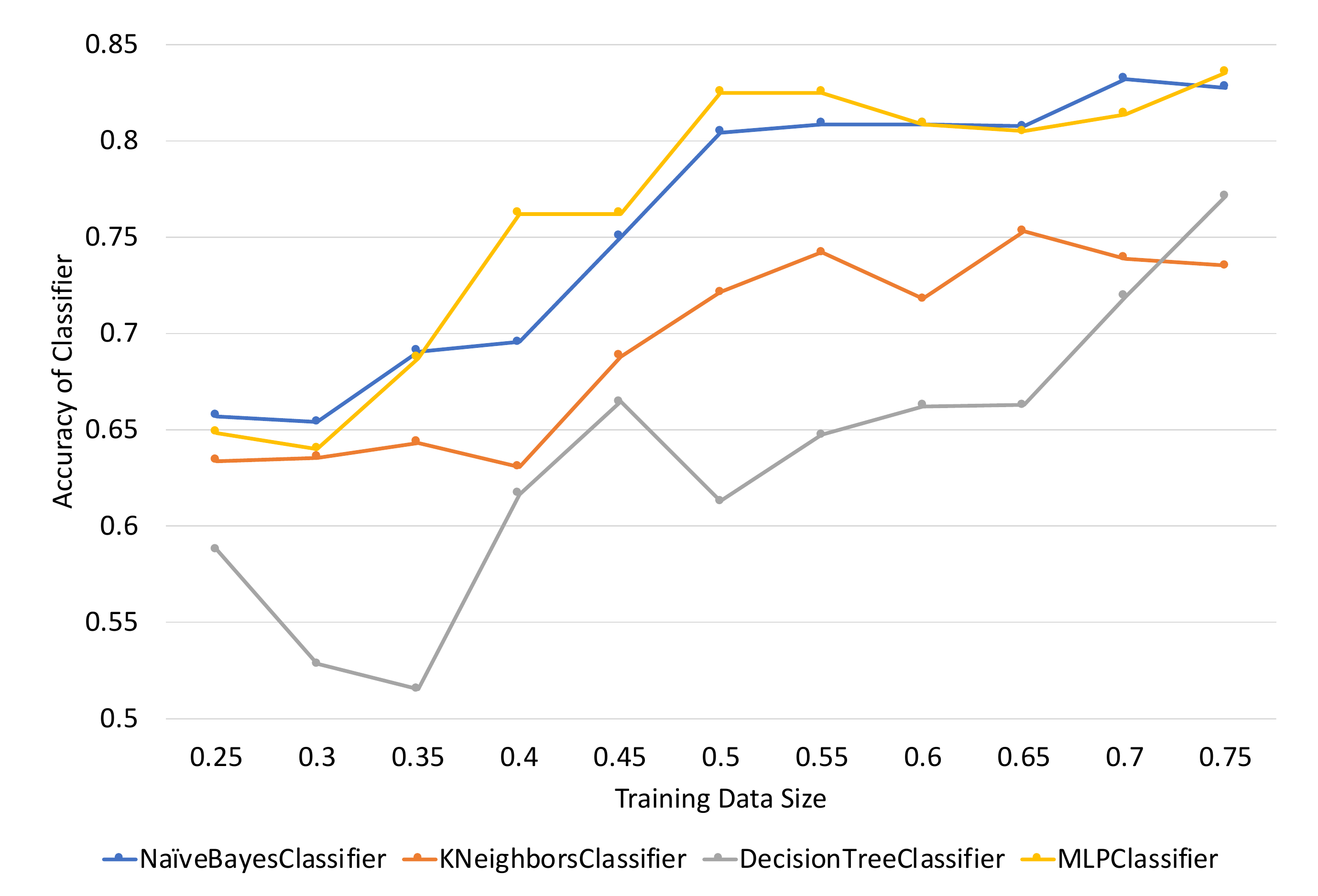}}
	\caption{Accuracy Comparison of Classifiers.}
	\label{fig5}
\end{figure}

\section{Conclusion}
We have build a sentiment analyzer to identify users' sentiment from Twitter hashtag \textit{\#kpujangancurang} toward the General Election Commission We use the hashtag to obtain a set of data from Twitter to analyse and investigate further the positive and the negative sentiment of the users from their tweets. This research utilizes rapid miner tool to generate the twitter data and comparing Naive Bayes, K-Nearest Neighbor, Decision Tree, and Multi-Layer Perceptron classification methods to classify the sentiment of the twitter data. There are overall 200 labeled data in this experiment. Overall, Naive Bayes and Multi-Layer Perceptron classifier outperformed the other two methods on 11 experiments with different size of training-testing data split. The two classifiers are potential to be used in creating sentiment analyzer for low-resource languages with small corpus. In our future work, we will compare the accuracy of both Naive Bayes and Multi-Layer Perceptron classifier on bigger size of corpus.

\section*{Acknowledgment}
This research is funded by Universitas Islam Riau.

\bibliographystyle{splncs04}
\bibliography{smartcyber}

\end{document}